\documentclass{article}
\usepackage{arxiv}

\usepackage{graphicx}

\usepackage{booktabs} 
\usepackage{nicefrac}  
\usepackage{microtype}
\usepackage[utf8]{inputenc}

\usepackage[dvipsnames]{xcolor}
\usepackage{wrapfig}
\usepackage{multirow}

\usepackage{amssymb}
\usepackage{bbm}
\usepackage{array}
\usepackage[colorlinks=true,urlcolor=purple,
linkcolor=purple,citecolor=purple]{hyperref}
\usepackage{amsmath}
\usepackage{mathtools}
\usepackage{float}

\usepackage{mathrsfs}
\usepackage{graphicx}
\usepackage{color}

\usepackage{algorithm}
\usepackage{algpseudocode}
\usepackage{thmtools,thm-restate} %

\makeatletter
\def\paragraph{\@startsection{paragraph}{4}%
  \z@\z@{-\fontdimen2\font}%
  {\normalfont\bfseries}}
\makeatother

\usepackage{tikz}
\usetikzlibrary{decorations.markings, backgrounds, calc, positioning, shapes, shadows, arrows, fit, automata}
\usepackage{tikz-cd}
\usetikzlibrary{arrows.meta}
\tikzset{>={Latex}}

\newcommand{\R}{\mathbb{R}}

\newcommand{\e}{\varepsilon}

\newcounter{desccount}

\newcommand{\descref}[1]{\hyperref[#1]{#1}}

\newcommand{\dgw}{d_{\textrm{GW}}}		%

\newcommand{\coup}{\mathscr{C}}

\newcommand{\X}{\mathbb{X}}

\makeatletter
\newcommand{\pushright}[1]{\ifmeasuring@#1\else\omit\hfill$\displaystyle#1$\fi\ignorespaces}
\newcommand{\pushleft}[1]{\ifmeasuring@#1\else\omit$\displaystyle#1$\hfill\fi\ignorespaces}
\makeatother

\usepackage{array}
\newcolumntype{P}[1]{>{\centering\arraybackslash}p{#1}}

\newcommand{\mcP}{\mathcal{P}}
\newcommand{\mcC}{\mathcal{C}}
\newcommand{\dqgw}{d_{\mathrm{qGW}}}

\newtheorem{theorem}{Theorem}
\newtheorem{proposition}[theorem]{Proposition}
\newtheorem{lemma}[theorem]{Lemma}

\begin{document}
\title{Quantized Gromov-Wasserstein}

\author{
  Samir Chowdhury\\
  Stanford University\\
     \And 
  David Miller \\
  University of Utah \\
  \And
 Tom Needham \\
  Florida State University\\
}

\maketitle              %
\begin{abstract}
The Gromov-Wasserstein (GW) framework adapts ideas from optimal transport to allow for the comparison of probability distributions defined on different metric spaces. Scalable computation of GW distances and associated matchings on graphs and point clouds have recently been made possible by state-of-the-art algorithms such as S-GWL and MREC. Each of these algorithmic breakthroughs relies on decomposing the underlying spaces into parts and performing matchings on these parts, adding recursion as needed. While very successful in practice, theoretical guarantees on such methods are limited. Inspired by recent advances in the theory of quantization for metric measure spaces, we define Quantized Gromov Wasserstein (qGW): a metric that treats parts as fundamental objects and fits into a hierarchy of theoretical upper bounds for the GW problem. This formulation motivates a new algorithm for approximating optimal GW matchings which yields algorithmic speedups and reductions in memory complexity. Consequently, we are able to go beyond outperforming state-of-the-art and apply GW matching at scales that are an order of magnitude larger than in the existing literature, including datasets containing over 1M points.

\end{abstract}

\section{Introduction}

It is frequently convenient to represent geometric data, such as the set of points in a point cloud or the set of nodes in a network, as a finite metric space. Such a representation naturally enjoys invariances to symmetries such as permutations or rigid Euclidean motions, which typically serve as nuisances in data analysis tasks. The well known Gromov-Hausdorff (GH) distance provides a mathematical framework for comparing metric spaces, but it is hard to handle computationally, as it inherently requires a \emph{point correspondence} between the spaces \cite{dgh-sgp,dgh-focm}. Gromov-Wasserstein (GW) distance \cite{dgh-sm, dghlp-focm} is a relaxation of Gromov-Hausdorff distance that compares \emph{metric measure (mm) spaces}---metric spaces endowed with probability measures---by optimizing a nonconvex loss over a convex domain. We precisely define GW distance below, but the idea is to compare mm-spaces by finding a \emph{probabilistic correspondence} (also referred to as a \emph{matching}) between their points. Matchings can be approximated via standard optimization tools in this relaxed setting.

While exact computation of GW distance is NP-Hard, recent algorithmic advances toward estimating it have made GW distance a viable tool for machine learning tasks on metric space-valued data at an increasingly large scale \cite{blumberg2020mrec,pcs16,xu2019scalable}. A common theme in recent approximation algorithms is to break the GW problem into smaller subproblems. A rough template for using this approach to compare two large mm-spaces $X$ and $Y$ is as follows: partition $X$ and $Y$ into smaller blocks of manageable size, find a matching between representatives of the blocks, then recurse this process to find matchings between the paired blocks. This recursive approach to estimating GW distance increases the feasible scale of $X$ and $Y$ by an order of magnitude ($\sim$1K points to $\sim$10K points). In this paper, we treat this partitioned matching paradigm more formally: we develop a theoretical formulation inspired by the duality between quantization and clustering developed recently in \cite{memoli2018sketching}, which in turn suggests a new approximation algorithm that scales to metric spaces with $\sim$100K or even $\sim$1M points. Our main contributions (code available at\footnote{\url{https://github.com/trneedham/QuantizedGromovWasserstein}}) are:
\begin{enumerate}
    \item we define a new metric on the space of partitioned mm-spaces called \emph{quantized Gromov-Wasserstein (qGW) distance};
    \item we give a novel algorithm for estimating qGW distance (Section \ref{sec:qGW}) and show that (for certain parameter choices) its complexity is \emph{nearly linear} in the sizes of the spaces being compared (Proposition \ref{prop:locally_linear_solution});
    \item we give theoretical error bounds comparing qGW and GW (Theorems \ref{thm:gw_estimation} and \ref{thm:delta_bound})---these error bounds give heuristics for the types of spaces on which qGW should be expected to perform well;
    \item we demonstrate empirically that qGW gives state-of-the-art performance for matching large scale shape datasets (Section \ref{sec:experiments}).
\end{enumerate}

\section{Quantized Gromov-Wasserstein: Theory and Computation}

\subsection{Distances Between Metric Measure Spaces}

\paragraph{Gromov-Wasserstein Distance.} A \emph{metric measure space}, or \emph{mm-space} for short, is a triple $(X,d_X,\mu_X)$ consisting of a compact metric space $(X,d_X)$ endowed with a Borel probability measure $\mu_X$. We abuse notation and abbreviate the triple simply as $X$. In practice, we deal with finite mm-spaces $X = \{x_1,\ldots,x_n\}$, where the metric can be represented as the distance matrix $(d_X(x_i,x_j))_{i,j}$ and the probability measure can be represented as a vector $(\mu_X(x_1),\ldots,\mu_X(x_n))$. Let $\coup(\mu_X,\mu_Y)$ denote the set of \emph{couplings} of $\mu_X$ and $\mu_Y$; i.e., Borel probability measures $\mu$ on the product space $X \times Y$ whose marginals are $\mu_X$ and $\mu_Y$. 

We define the \emph{Gromov-Wasserstein loss} of a coupling $\mu \in \mathcal{C}(\mu_X,\mu_Y)$ as
\begin{equation}\label{eqn:gw_loss}
    \mathrm{GW}(\mu) = \iint_{(X\times Y)^2} (d_X(x,x')-d_Y(y,y'))^2 \, \mu(dx,dy) \mu(dx',dy').
\end{equation}
The function $\mathrm{GW}$ depends on the mm-spaces $X$ and $Y$, but we suppress this dependence from the notation for convenience. In the finite mm-space setting, which will be the focus of this paper, \eqref{eqn:gw_loss} can be expressed as 
\begin{equation}\label{eqn:finite_gw_loss}
    \mathrm{GW}(\mu) = \sum_{i,j,k,\ell} (d_X(x_i,x_k) - d_Y(y_j,y_\ell))^2 \mu(x_i,y_j) \mu(x_k,y_\ell).
\end{equation}
The \emph{Gromov-Wasserstein distance} between $X$ and $Y$ is 
\begin{equation}\label{eqn:gw_distance}
\dgw(X,Y) := \inf_{\mu \in \mathcal{C}(\mu_X,\mu_Y)} \mathrm{GW}(\mu)^{1/2}.
\end{equation}
For finite spaces, $\mathcal{C}(\mu_X,\mu_Y) \subset \R^{|X| \times |Y|}$ is a convex polytope whose elements are represented as matrices. In this setting, a minimizer $\mu$ of \eqref{eqn:finite_gw_loss} can be considered as a \emph{soft alignment} of the spaces $X$ and $Y$: the row of $\mu$ (considered as a matrix) indexed by $x_i$ gives a probabilistic assignment to each of the points of $Y$. We frequently refer to an optimizer of \eqref{eqn:finite_gw_loss} as a \emph{matching} of the mm-spaces.

We remark that Equation \ref{eqn:gw_distance} is one of two different notion of Gromov-Wasserstein distance \cite{dgh-sm,dghlp-focm,sturm2006geometry} in the literature, and the main theoretical properties of both these Gromov-Wasserstein distances were worked out by M\'{e}moli \cite{memoli2008gromov,dgh-sm,dghlp-focm} and Sturm \cite{sturm2006geometry,sturm2012space}. The formulation in Equation (\ref{eqn:gw_distance}) is the one used more often in computational settings \cite{peyre2019computational}, and we focus on this case throughout the current work. After initial theoretical development, large scale applications of GW to machine learning and graphics problems were later explored in \cite{pcs16,solomon2016entropic}, and a vast literature on GW distance has since developed, focusing on both theoretical \cite{memoli2018gromov,gwnets,chowdhury2021generalized,chowdhury2020gromov,sejourne2020unbalanced,weitkamp2020gromov} and applications-driven \cite{alvarez2018gromov,bunne2019learning,demetci2020gromov} aspects.

For finite mm-spaces $X$ and $Y$ of size $\approx n$,  naive evaluation of the objective function  \eqref{eqn:finite_gw_loss} incurs cost $O(n^4)$. A less naive implentation brings this down to $O(n^3 \log(n))$ \cite{pcs16}, but this is still prohibitively expensive for even medium scale tasks with on the order of several thousand points. In Section \ref{sec:qGW_metric}, we introduce a metric which operates on partitioned spaces, thereby formalizing ideas which have appeared in more ad hoc forms in the recent literature \cite{vayer2019sliced,xu2019scalable,blumberg2020mrec}; we make connections to these methods precise in Section \ref{sec:related_work}. An efficient algorithm for approximating this metric is presented in Section \ref{sec:qGW}.

\paragraph{The Quantized Gromov-Wasserstein Metric.}\label{sec:qGW_metric}

Recent approaches to approximating GW distance have used a divide-and-conquer strategy, where the spaces are partitioned into blocks and the blocks are matched recursively. In this section, we develop a formal framework for matching partitioned spaces. We treat a partition of a space as input data for a metric, obtained in a preprocessing step; practical methods for finding good partitions are described in Section \ref{sec:qGW}.

Let $X$ be a finite mm-space. A \emph{pointed partition} of $X$ is a structure consisting of a partition of $X$ into disjoint, nonempty sets $U^1,\ldots,U^m$ together with a representative point $x^p \in U^p$ for each $p = 1,\ldots,m$. We denote this structure as $\mathcal{P}_X = \{(x^1,U^1),\ldots,(x^m,U^m)\}$. When a pointed partition has $m$ points, we will refer to it as an \emph{$m$-pointed partition}. We will typically work in the regime $m \ll |X|$. A set $U^p$ in $\mathcal{P}_X$ will be referred to as a \emph{partition block} and the distinguished point $x^p \in U^p$ will be referred to as the \emph{representative} of the partition block. Note that we use superscripts $x^p$ to denote partition block representatives, to distinguish from generic indexed points $x_i$ in the metric space. An \emph{$m$-pointed metric measure space} is a quadruple $(X,d_X,\mu_X,\mathcal{P}_X)$; this notation will be abbreviated as $(X,\mathcal{P}_X)$ when the existence of a chosen metric and measure is clear from context. 

To an $m$-pointed mm-space $(X,\mathcal{P}_X)$ we associate several related mm-spaces. Let $X^m := \{x^1,\ldots,x^m\}$ denote the set of partition block representatives. There is a  well-defined projection map $X \to X^m$ induced by the partition, i.e. the map $x\mapsto x^p$ when $x \in U^p$. We endow $X^m$ with the measure $\mu_{\mcP_X}$  given by the pushforward of $\mu_X$ by this projection map. Then $X^m$ has an mm-space structure given by restricting $d_X$. We refer to $X^m = (X^m,d_X|_{X^m},\mu_{\mcP_X})$ as a \emph{quantized representation} of $X$. For each $(x^p,U^p) \in \mathcal{P}_{X}$, we also obtain a new mm-space $(U^p,d_X|_{U^p},\mu_{U^p})$, where 
\[
\mu_{U^p} := \frac{1}{\mu_X(U^p)} \mu_X|_{U^p}.
\]
This can be extended to a probability measure $\bar{\mu}_{U^p}$ on all of $X$ via the formula 
\begin{equation}\label{eqn:extending_measure}
\bar{\mu}_{U^p}(A) := \mu_{U^p}(A \cap U^p) \qquad \forall \; A \subset X.
\end{equation}

Let $(Y,d_Y,\mu_Y,\mcP_Y)$ be another finite $m$-pointed mm-space with quantized representation $Y^m$. A \emph{quantization coupling} is a measure $\mu$ on $X \times Y$ of the form
\begin{equation}\label{eqn:quantization_coupling}
\mu(x,y) = \sum_{p,q} \mu_m(x^p,y^q) \bar{\mu}_{x^p,y^q}(x,y),
\end{equation}
where $\mu_m \in \mcC(\mu_{\mcP_X},\mu_{\mcP_Y})$, each $\mu_{x^p,y^q} \in \mcC(\mu_{U^{p}},\mu_{V^{q}})$ and $\bar{\mu}_{x^p,y^q} \in \mcC(\bar{\mu}_{U^p},\bar{\mu}_{V^q})$ is an extension of $\mu_{x^p,y^q}$ using the same trick as \eqref{eqn:extending_measure}. We moreover assume that $\mu_{x^p,y^q}(x^p,y^q) > 0$ for each $p,q$. Let $\mathcal{C}_{\mcP_X,\mcP_Y}(\mu_X,\mu_Y)$ denote the set of quantization couplings of $\mu_X$ and $\mu_Y$ with respect to $\mcP_X$ and $\mcP_Y$. We have the following proposition, which says that quantization couplings are couplings in the usual sense. The proof is provided in the supplementary materials.

\begin{proposition}\label{prop:quantization_couplings}
    Any quantization coupling is a coupling; that is, the set of quantization couplings $\mathcal{C}_{\mcP_X,\mcP_Y}(\mu_X,\mu_Y)$ is a subset of the set of couplings $\mathcal{C}(\mu_X,\mu_Y)$.
\end{proposition}

We define the \emph{quantized Gromov-Wasserstein (qGW) distance} between finite $m$-pointed mm-spaces as
\begin{equation}\label{eqn:quantized_gw}
    \dqgw((X,\mcP_X),(Y,\mcP_Y)) := \inf_{\mu \in \mathcal{C}_{\mcP_X,\mcP_Y}(\mu_X,\mu_Y)} \mathrm{GW}(\mu)^{1/2}.
\end{equation}

An \emph{isomorphism} of mm-spaces $X$ and $Y$ is a measure-preserving isometry $X \to Y$. Let $\mathcal{M}$ denote the collection of all isomorphism classes of finite mm-spaces. This notion can be specialized to pointed mm-spaces: an \emph{isomorphism} of $m$-pointed mm-spaces $(X,\mcP_X)$ and $(Y,\mcP_Y)$ is an isomorphism $X \to Y$ which takes $X^m$ to $Y^m$. We use $\mathcal{M}^m$ to denote the collection of all finite $m$-pointed mm-spaces, considered up to isomorphism.

\begin{theorem}\label{thm:qGW_is_a_metric}
    The quantized GW distance $\dqgw$ is a metric on $\mathcal{M}^m$.
\end{theorem}

The interesting part of the proof is establishing the triangle inequality, which is an application of the gluing lemma from optimal transport theory \cite[Lemma 7.6]{villani2003topics}. The idea is to produce a coupling of $(X,\mcP_X)$ and $(Z,\mcP_Z)$ by applying the gluing lemma to quantization couplings of $(X,\mcP_X)$ and $(Y,\mcP_Y)$ and $(Y,\mcP_Y)$ and $(Z,\mcP_Z)$. A lengthy computation shows that the result is automatically a quantization coupling, and the triangle inequality follows. A full proof of the theorem is provided in the supplementary materials.

Exact computation of quantized GW distance $\dqgw$ is intractable, but the structure of the metrics suggests natural heuristics. We give an algorithm for approximating $\dqgw$ in Section \ref{sec:qGW}. Theoretical estimates for the quality of this approximation with respect to the standard GW distance between underlying mm-spaces are presented in Section \ref{sec:error_bounds}.

\subsection{The Quantized Gromov-Wasserstein Algorithm}\label{sec:qGW}

Let $(X,d_X,\mu_X,\mcP_X)$ and $(Y,d_Y,\mu_Y,\mcP_Y)$ be $m$-pointed mm-spaces and let $X^m$ and $Y^m$ denote their quantized representations. In this subsection, we present an efficient method for approximating quantized GW distance $\dqgw$ between these spaces, which in turn gives an estimate of GW distance between the underlying mm-spaces. The algorithm proceeds in three steps:

\begin{enumerate}
    \item {\bf Global Alignment:} The first step in the algorithm is to compute $\mu_m$, an optimal coupling of the quantized representations of $X$ and $Y$---$m \ll |X|,|Y|$ is chosen so that this optimal coupling can be feasibly approximated via existing methods, such as those implemented in the Python Optimal Transport library \cite{flamary2017pot}. This serves to give a \emph{global alignment} of the spaces. 
    
    \item {\bf Local Alignment:} The second step is to produce a collection of \emph{local alignments}. For each $x^p \in X^m$ and $y^q \in Y^m$, we obtain a coupling $\mu_{x^p,y^q}$ of the partition block mm-spaces $(U^p,d_X|_{U^p},\mu_{U^p})$ and $(V^q,d_Y|_{V^q},\mu_{V^q})$ by solving the optimal transport problem
\begin{equation}\label{eqn:locally_linear}
    \min_{\mu_{x^p,y^q} \in \mathcal{C}(\mu_{U^p},\mu_{V^q})} \sum_{x \in U^p, y \in V^q} (d_X(x,x^p) - d_Y(y,y^q))^2 \mu_{x^p,y^q}(x,y).
\end{equation}
The solution $\mu_{x^p,y^q}$ is referred to as a \emph{local linear matching} of $U_p$ and $V_q$. The simplified matching problem \eqref{eqn:locally_linear} can be solved efficiently, as it is equivalent to finding an optimal transport plan between distributions on the real line---this is made precise in Proposition \ref{prop:locally_linear_solution} below. 

\item {\bf Create Coupling:} The final step is to create a quantization coupling from the global and local alignments,
\[
\mu = \sum_{p,q} \mu_m(x^p,y^q)\overline{\mu}_{x^p,y^q}.
\]
This is treated as an approximate solution to the $\dqgw$ optimization problem. 
\end{enumerate}

We remark that the local linear matching obtained by solving \eqref{eqn:locally_linear} is generally \emph{not} a solution of the GW optimization problem \eqref{eqn:gw_distance}, applied to the partition blocks $U^p$ and $V^q$. An alternative approach to approximating $\dqgw$ would be to replace the local alignment step above with a local alignment which actually solves the GW subproblems---indeed, this procedure is similar to what is done in the Scalable Gromov-Wasserstein \cite{xu2019gromov} and MREC \cite{blumberg2020mrec} frameworks. We show in Section \ref{sec:computational_complexity} that our simplified algorithm drastically reduces the computational complexity over computing several GW subproblems.

A subroutine in this procedure is a heuristic for generating good partitions. In the graph setting, we applied the Fluid community detection algorithm \cite{pares2017fluid}---available in the Python package \texttt{networkx} \cite{hagberg2008exploring}---to choose partition blocks, and we chose block representatives with maximal PageRank \cite{brin1998anatomy}. In the point cloud settings we simply chose uniform iid samples without replacement and computed a Voronoi partition; more principled approaches such as $k$-means and its variants are of course possible.

\paragraph{Computational Complexity.}\label{sec:computational_complexity}

We now give bounds on the computational complexity of finding a quantization coupling via local linear matchings, as was described in Section \ref{sec:qGW}. The key observation is that the optimization problem \eqref{eqn:locally_linear} can be solved extremely efficiently. The next proposition follows from a result on pushforward measures \cite[Lemma 27]{gwnets} and from the well known fact that one-dimensional optimal transport can be solved efficiently \cite[Section 2.6]{peyre2019computational}; see the supplementary material for details.

\begin{proposition}\label{prop:locally_linear_solution}
    The optimization problem \eqref{eqn:locally_linear} is equivalent to a one dimensional optimal transport problem and can therefore be solved in $O(k\log(k))$ time, with $k$ the max number of points in $U^p$ or $V^q$.
\end{proposition}

We can therefore estimate the computational complexity of the qGW approximation algorithm as follows. Suppose that $|X|$ and $|Y|$ are of order $N$ and that they are each partitioned into $m$ blocks. Also suppose that the blocks are of roughly equal size, so that there are approximately $N/m$ points per block. The worst-case complexity of the quantized GW algorithm is the maximum of an iterative $O(m^3 \log(m))$ term coming from approximation of the global GW alignment (via some variant of gradient descent) \cite{pcs16} and a $O(m^2 \cdot N/m \log (N/m))$ term coming from performing $m^2$ local linear matchings. However, it has been observed empirically \cite{xu2019scalable,chowdhury2020gromov} that optimal GW couplings tend to have supports whose sizes scale \emph{linearly} in the number points in the spaces being matched (rather than the worst-case quadratic scaling)---in fact, this order of scaling can be theoretically guaranteed when comparing symmetric positive definite kernel matrices \cite{chowdhury2021generalized}. Since local linear matchings only need to be computed for $x^p,y^q$ such that $\mu_m(x^p,y^q) \neq 0$, the expected computational complexity for the algorithm is therefore the maximum of an iterative $O(m^3 \log(m))$ term and an $O(N \log (N/m))$ term; taking $m$ on the order of $N^{1/3}$, for example, gives an algorithm with computational cost $O(N\log(N))$ (iterative).

Memory complexity of GW computations becomes a serious issue at scale. For example, storing and manipulating distance matrices on 50K points (as in the David mesh, cf. the graph matching experiment in \S\ref{sec:exp-graph-match}) with 64-bit floats requires 20GB memory. We resolve this by observing that qGW never requires the full $O(N^2)$ distance matrices: we only require storing a dense $O(m^2)$ matrix of distances between representatives, and a sparse $O(Nm)$ matrix of distances between each block representative to the points \emph{in the same block}. In addition to enabling qGW computations on datasets with $\sim$1M points (cf. the large scale matching experiment in \S\ref{sec:exp-large-matching}), this observation becomes especially useful when preprocessing distance matrices on graphs: instead of incurring $O(N|E|\log(N))$ cost of computing a full matrix of graph geodesic distances via Dijkstra's algorithm, we simply incur $O(m|E|\log(N))$ cost. Here $E$ denotes edges.

Finally, the quantization approach allows for fast computation of individual queries, i.e. individual rows of the coupling matrix. Given a point $x\in X$ with block representative $x^p$, we can compute $\mu(x,\cdot)$ (i.e. the target of $x$ in $Y$) by only accessing the $m^2$ matrices of distances between representatives in $X,Y$, the distances from $x^p$ to all other points in its block, and likewise for all $y^q$ such that $\mu_m(x^p,y^q)>0$ (typically $\mu_m$ is sparse).

\subsection{Quantized Fused Gromov-Wasserstein}

It is common that data, represented as a finite metric space $(X,d_X)$, comes endowed with attributions; i.e., with a function $f:X \to Z$ valued in another metric space $(Z,d_Z)$. This is the case, for example, when $X$ represents (the nodes of a) network and $f:X \to Z$ represents node attributes, which can be data-driven or crafted from local network features. This extra complexity is handled elegantly in the GW framework by the Fused Gromov-Wasserstein (FGW) distance of Vayer et al.\ \cite{vayer2019optimal}. We give a brief formulation of FGW distance in the setting of finite metric spaces.

Let $(Z,d_Z)$ be a metric space. A \emph{finite $Z$-structured mm-space} is a quadruple $(X,d_X,\mu_X,f_X)$, where $(X,d_X,\mu_X)$ is a finite mm-space and $f_X:X \to Z$ is an arbitrary function. We denote this structure as $(X,f_X)$ when the existence of a metric and measure on $X$ is clear from context. Let $(X,f_X)$ and $(Y,f_Y)$ be $Z$-structured mm-spaces and let $\alpha > 0$. The \emph{Fused Gromov-Wasserstein loss}, with parameter $\alpha$, of a coupling $\mu \in \mathcal{C}(\mu_X,\mu_Y)$ is 
\[
\mathrm{FGW}_\alpha(\mu) := (1-\alpha) \mathrm{GW}(\mu) + \alpha \mathrm{W}(\mu),
\]
where
\[
\mathrm{W}(\mu) := \sum_{i,j} d_Z(f_X(x_i),f_Y(y_j))^2 \mu(x_i,y_j)
\]
is classical Wasserstein loss. One then defines \emph{FGW distance} as
\[
d_{\mathrm{FGW},\alpha}((X,f_X),(Y,d_Y)) := \min_{\mu \in \mathcal{C}(\mu_X,\mu_Y)} \mathrm{FGW}_\alpha(\mu)^{1/2}.
\]

We now describe a quantized algorithm for approximating FGW distance. Let $(X,d_X,\mu_X,\mathcal{P}_X,f_X)$ and $(Y,d_Y,\mu_Y,\mathcal{P}_Y,f_Y)$ be $m$-partitioned mm-spaces endowed with $Z$-structures. The approximation algorithm proceeds in steps similar to the qGW algorithm, and involves an additional parameter $\beta$.

\begin{enumerate}
    \item {\bf Global Alignment:} The first step in the approximation algorithm is to determine a global registration. This is done by computing a coupling $\mu_m \in \mathcal{C}(\mu_X,\mu_Y)$ as a minimizer of $\mathrm{FGW}_\alpha$ for the $Z$-structured mm-spaces $(X^m,d_X|_{X^m},\mu_{\mathcal{P}_X},f_X|_{X^m})$ and $(Y^m,d_Y|_{Y^m},\mu_{\mathcal{P}_Y},f_Y|_{Y^m})$.
    \item {\bf Local Alignment:} Next we find a $\mu_{x^p,y^q} \in \mathcal{C}(\mu_{U^p},\mu_{V^q})$ for each $x^p,y^q$. We first solve the local linear matching problem \eqref{eqn:locally_linear} to get a coupling $\mu_{x^p,y^q}^{(0)}$. Next, we solve another local linear matching problem with respect to $Z$-valued features
    to obtain a second coupling $\mu_{x^p,y^q}^{(1)}$. Finally, we define $\mu_{x^p,y^q}$ by a simple weighted average with respect to our parameter $\beta$, and set
    \[
    \mu_{x^p,y^q} = (1-\beta)\mu^{(0)}_{x^p,y^q} + \beta \mu^{(1)}_{x^p,y^q}.
    \]
    \item {\bf Create Coupling:} We create a coupling, as in the qGW algorithm.
\end{enumerate}

The parameters in this matching algorithm can therefore be described intuitively: $\alpha$ controls the preference to globally match based on metric structure or feature structure, while $\beta$ controls this preference locally.

\subsection{Related Work}\label{sec:related_work}

We now describe in more detail other approaches to scaling the GW framework and the relationship between our algorithm and others in the literature. The first serious attempt to scaling GW distance computation is the \emph{scalable Gromov-Wasserstein (sGW)} framework of Xu et al. \cite{xu2019scalable}. This framework is designed specifically to generate GW matchings between graphs, represented as adjacency matrices, by leveraging the observations of \cite{gwnets,pcs16} that the GW framework gives a sensible way to compare arbitrary square matrices (not just distance matrices). Xu et al. introduced a recursive partitioning scheme, allowing a ``divide-and-conquer" approach to the matching problem. A similar approach to approximating GW matchings is the basis of the MREC algorithm of Blumberg et al. \cite{blumberg2020mrec}, which finds matchings between Euclidean point clouds or more general metric spaces via a scheme which recursively partitions the data and defines smaller subproblems by matching partition block representatives. The MREC framework is quite general, also allowing matchings based on classical Wasserstein distance. Our algorithm qGW fits into the general mold of sGW and MREC, but we replace the recursive definition of submatching problems with a simpler local linear matching problem. More broadly, these methods based on partitioning are related, at least in the metric space setting, to a duality between quantization and clustering as studied in \cite{memoli2018sketching}. We use this connection explicitly when obtaining error bounds for qGW. 

The ideas presented in this paper are related to other common themes appearing in the GW and broader optimal transport literature. Recently introduced by Vayer et al. as a fast approximation of GW distance, the \emph{sliced Gromov-Wasserstein distance} \cite{vayer2019sliced} computes a dissimilarity between Euclidean point clouds by taking an expectation of Gromov-Wasserstein distance between random 1-dimensional projections. By its nature, this method is limited to Euclidean point clouds, but extra optimization steps can be incorporated to compare point clouds lying in different dimensions and to make the dissimilarity rotation-invariant. Our qGW algorithm also speeds up the GW computation by using 1-dimensional projections; here, we are invoking a 1-dimensional problem by ``slicing" radially from prealigned anchor points. This approach to slicing means that our algorithm works on general metric spaces and that it is naturally invariant to isometries such as rigid motions. The idea of using distances to anchor points to compare mm-spaces is used in computable \emph{lower} bounds on GW distance derived in \cite{dghlp-focm} (whereas our method always produces an \emph{upper} bound). We remark that efficient algorithms for computing variants of this lower bound have recently been introduced by Sato et al.\ \cite{sato2020fast}. Although not directly related to this work, similar ideas involving finding optimal anchor points for simplified graph representations via methods of optimal transport have appeared in \cite{garg2019solving}.

\section{Theoretical Error Bounds}\label{sec:error_bounds}

\paragraph{Quantized Eccentricity.}

Let $(X,\mcP_X)$ be an $m$-pointed mm-space. As above, we let $X^m = \{x^1,\ldots,x^m\}$ denote the set of partition block representatives. We treat $X^m$ and each of the blocks $U^p$ as a mm-space, as described in Section \ref{sec:qGW_metric}. The goal of this section is to obtain estimates on the approximation quality of the qGW algorithm with respect to the true GW distance.

Let $x \in X$. The \emph{eccentricity} \cite{dghlp-focm} of $x$ is 
\[
s_X(x):= \left(\sum_{x'} d_X(x,x')^2 \mu_X(x')\right)^{1/2}.
\]
We define the \emph{quantized eccentricity} of $\mcP_X$ as 
\[
q(\mcP_X) := \left(\sum_p \mu_X(U^p) s_{U^p}(x^p)^2\right)^{1/2}.
\]
The \emph{$m$-quantized eccentricity} of $X$, denoted $q_m(X)$, is the minimum $q(\mcP_X)$ over all $m$-pointed partitions of $X$.

\begin{lemma}\label{lem:quantized eccentricity_bound}
For a finite mm-space $X$ and $m$-pointed partition $\mcP_X$ with partition block representatives $X^m$, 
\[
\dgw(X,X^m) \leq 2 q(\mcP_X).
\]
It follows that
\[
\min_{\mcP_X} \dgw(X,X^m) \leq 2 q_m(X),
\]
where the minimum is over $m$-pointed partitions $\mcP_X$.
\end{lemma}

Intuitively, the lemma says that a mm-space with small $m$-quantized eccentricity is well-approximated by an $m$-point subset. The proof is given in the supplementary materials. The strategy of the proof is inspired by the proof of \cite[Theorem 1.12]{memoli2018sketching}, which compares other abstract measures (related to quantization and clustering) for coarsely representing a mm-space. Combining the lemma with the reverse triangle inequality immediately yields the following result.

\begin{theorem}\label{thm:gw_estimation}
For finite mm-spaces $X$ and $Y$,
\begin{equation}\label{eqn:quantized eccentricity_bound}
\min_{\mcP_X,\mcP_Y} \left|\dgw(X,Y) - \dgw(X^m,Y^m) \right| \leq 2\left(q_m(X) + q_m(Y)\right).
\end{equation}
where the minimum is taken over $m$-pointed partitions $\mcP_X$ and $\mcP_Y$. 
\end{theorem}

We note that neither side of the estimate \eqref{eqn:quantized eccentricity_bound} is explicitly computable. The theorem should be interpreted as giving intuition about which types of mm-spaces are amenable to accurate comparison by the qGW algorithm. Since the qGW algorithm described above begins by finding a global alignment of mm-spaces using partition block representatives, we would like to understand how well this global alignment reflects the true distance between mm-spaces. The result says that one can only reasonably hope for an accurate representation when the mm-spaces in question have low quantized eccentricity. As a heuristic, this is the case for dense point clouds in low dimensions or for graphs with rigid geometric structure. Low quantized eccentricity is less likely in high-dimensional point clouds due to the concentration of measure phenomenon or in graphs such as social networks, due to typical ``small world" structure.

\paragraph{Error bound for the qGW Algorithm.}

Let $(X,\mcP_X)$ and $(Y,\mcP_Y)$ be $m$-pointed mm-spaces and let
\[
\delta((X,\mcP_X),(Y,\mcP_Y)) := \mathrm{GW}(\mu)^{1/2},
\]
where $\mu$ is the coupling obtained by the qGW algorithm with locally linear matchings described in Section \ref{sec:qGW}.

\begin{theorem}\label{thm:delta_bound}
Let $(X,\mcP_X)$ and $(Y,\mcP_Y)$ be $m$-pointed mm-spaces such that the metric diameter of each partition block in both $\mcP_X$ and $\mcP_Y$ is bounded above by $\epsilon > 0$. Then
\[
\left|\dgw(X,Y) - \delta((X,\mcP_X),(Y,\mcP_Y)) \right| \leq 2\left(q(\mcP_X) + q(\mcP_Y)\right) + 8\epsilon. 
\]
\end{theorem}

The proof uses the quantized eccentricity results of the previous subsection together with some careful estimates on the $\delta$ dissimilarity; details are provided in the supplementary materials. The value of this result is that it once again gives a heuristic for when quantized GW will be most useful: when the spaces being compared admit partitions with small quantized energy and with partition blocks of small diameter.

\section{Experiments}\label{sec:experiments}

\paragraph{Point Cloud Matching.}\label{sec:point_cloud_matching}

In this experiment we investigate the power of several scalable variants of GW distance to uncover ``ground truth" matchings between 3D point clouds. We use point clouds from the CAPOD dataset \cite{papadakis2014canonically}, which contains several classes of 3D meshes (we use the vertices as point clouds) of various sizes. For each class, we choose 10 shape samples, each treated as a mm-space by endowing it with Euclidean distance and uniform measure. For each shape sample, we create a copy whose vertices are permuted and perturbed (randomly within 1\% of the diameter of the shape). The task is to use a GW matching to get correct matches between points in the original shape $X$ and its noisy, permuted copy $\widetilde{X}$. Given a matching $\mu$ (a probabilistic correspondence between points in $X$ and $\widetilde{X}$), we compute the \emph{distortion} for each point $x_i \in X$ as the distance from its ground truth copy $\widetilde{x}_i$ and its \emph{matched point} $y_j := \mathrm{argmax}_{y_j} \mu(x_i,y_j)$. The \emph{distortion score} for the matching is the mean squared distortion.

\begin{figure}
    \centering
    \includegraphics[width = 0.8\textwidth]{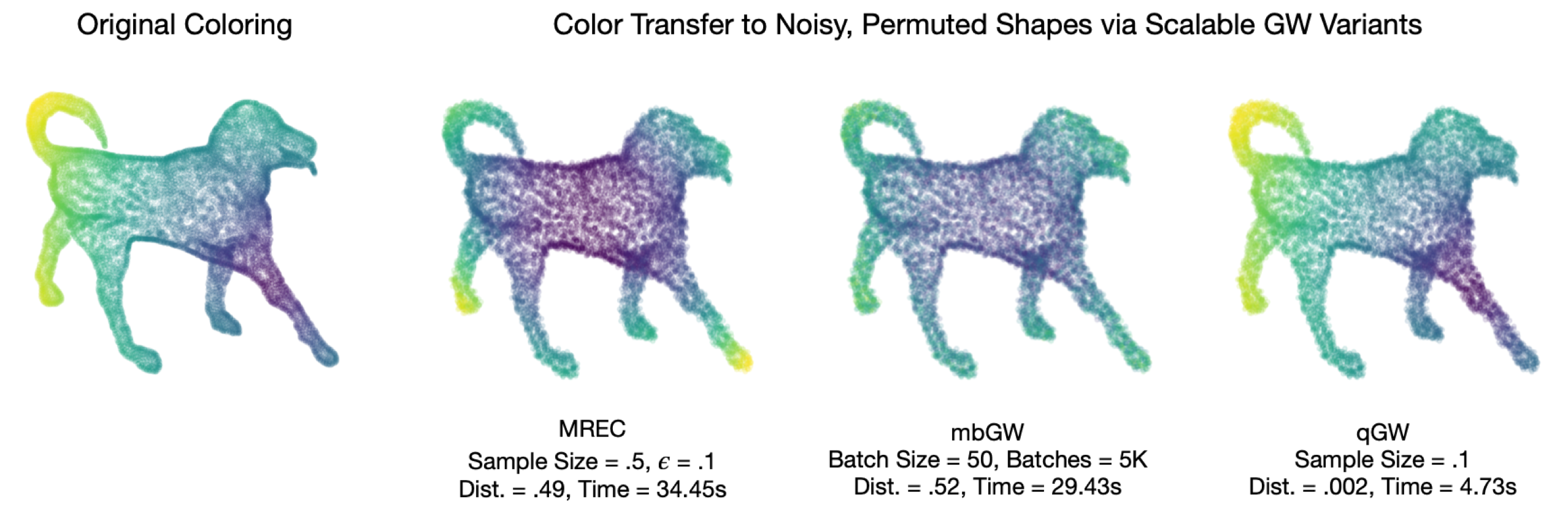}
    \caption{Visualizing point cloud matchings. We match the dog point cloud on the left ($\sim$9K points) to a copy whose points have been perturbed and whose order has been permuted. Matchings are computed via scalable variants of GW distance: MREC \cite{blumberg2020mrec}, minibatch GW \cite{fatras2021minibatch} and our qGW algorithm, with parameters as described in the figure. The distortion (see text) and compute time of each matching are provided. We visualize the matching by transferring a coloring of the points of the original shape to the new shapes via the respective matchings. Each algorithm provides a probabilistic correspondence between points; the color of a point in the target shape is a weighted average of the colors of the points in the source shape, with the weights given by the probability that a source shape matches to the query target shape.}
    \label{fig:color_transfer}
\end{figure}

For each class and each sample, matchings are produced by several algorithms; in Table \ref{tab:point_cloud_matching} we report class average distortion scores and compute times. We test the qGW algorithm, where we choose an $m$-pointed partition of each shape sample $X$ by randomly sampling $\lfloor p\cdot |X| \rfloor$ points, $p \in \{.01,.1,.2,.5\}$, as partition block representatives and then taking a  Voronoi partition with respect to these representatives. We also test against several baselines: standard GW is used as an overall baseline, where we see that compute times quickly become infeasible for any large scale task. For scalable baselines, we compare against entropy regularized GW (erGW) \cite{pcs16}, MREC \cite{blumberg2020mrec} and minibatch GW (mbGW) \cite{fatras2021minibatch}. Entropy regularized GW has a regularization weight parameter $\epsilon$ and we test $\epsilon \in \{0.2,5\}$ to check results in low and high regularization regimes. The MREC algorithm is very flexible and can incorporate several clustering and matching methods into its overall architecture. Our comparison does not reflect the full capabilities of MREC, and we only used parameters giving a direct comparison to other GW-based methods; that is, we use the GW module for matching and the random Voronoi partitioning  module for clustering. With these choices, MREC has two additional parameters $(\epsilon,p)$, where $\epsilon$ is a regularization weight parameter and $p$ is the percentage of points used when creating partitions for recursion. We used $\epsilon \in \{0.1,5\}$---we ran a larger grid search, and the reported results are qualitatively similar to those we obtained for other parameters---and $p \in \{.01,.1,.2,.5\}$. Minibatch GW has parameters $(n,k)$, where $n$ the number of samples per batch and $k$ is the number of batches---either an integer or a fraction of the size of each dataset. We use $n = 50$, and $k = 5K$ or $10\%$ of the size of the datasets (following the method of \cite[Figure 16]{fatras2021minibatch}). We remark that we are not aware of an official implementation for obtaining mbGW matchings, and the results here were created by our own implentation. Representative matchings for several methods are shown in Figure \ref{fig:color_transfer}.

\begin{table*}
    \centering
    \caption{Distortion scores (lower is better) and runtimes for variants of Gromov-Wasserstein matchings. The average number of points in each shape class is provided under the shape class name. Results are listed for several parameter choices of each method---see the text for details. Text in {\bf bold} is the best score across scalable GW methods and \underline{underlined} text is the the best score among the methods in the top 50\% of methods in terms of compute time. Blank entries correspond to experiments which failed to complete in 10 hours.}
    \label{tab:point_cloud_matching}
    \resizebox{\textwidth}{!}{
    \begin{tabular}{
    p{0.09\textwidth} p{0.09\textwidth} 
    P{0.15 \textwidth} P{0.15\textwidth}
    P{0.15\textwidth} P{0.15\textwidth}
    P{0.16\textwidth} P{0.18\textwidth} P{0.15\textwidth}}
    \toprule 
    Method & Param &
    Humans & Planes & Spiders & Cars
    & Dogs & Trees & Vases \\
    {} & {} &
    1926 & 2144 & 2664 & 5220
    & 8937 & 10433 & 15828 \\
    \midrule
    GW & {} & .07 (8.47) & .08 (19.64)  & .005 (28.29) & .16 (99.83) & .003 (512.80) & .015 (835.46) & --- \\
    \midrule
    \multirow[t]{4}{*}{erGW} & $0.2$ & {\bf.03 (15.42)} & .09 (17.03) & .090 (10.54)  & {\bf .16 (85.89)} & .153 (920.77) & .200 (2490.14) & --- \\
    {} & $5$ & .63 (2.93) & .87 (6.23) & .182 (9.76) & .67 (41.03) &.687 (181.94) & .714 (275.73) & .61 (1312.87) \\
    \midrule
    \multirow[t]{4}{*}{MREC} & $(.1,.01)$ & .32 (1.45) & .52 (2.37) & .060 (1.88) & .87 (3.12) & .459 (8.89) & .472 (9.05) & .43 (23.49) \\
    {} & $(5,.01)$ & .76 (.29) & .76 (.43) & .131 (0.52) & .98 (1.75) & .653 (5.31) &.549 (6.48) & .58 (15.66) \\
    {} & $(.1,.1)$ & .25 (2.31) & .39 (3.37) & .064 (2.61) & .38 (8.00) & .378 (15.59) & .316 (23.20) & .48 (43.05) \\
    {} & $(5,.1)$ & .67 (.54) & .83 (.82) & .166 (1.02) & .63 (2.94) & .674 (8.36) &.580 (10.87) & .62 (29.07) \\
    {} & $(.1,.2)$ & .32 (1.54) & .10 (2.42) & .065 (2.99) & .20 (11.24) & .415 (30.86) & .391 (40.58) & .50 (130.86)\\
    {} & $(5,.2)$ & .53 (1.31) & .89 (2.10) & .160 (2.80) & .72 (8.88) & .686 (26.97) & .720 (37.48) & .60 (100.51) \\
    {} & $(.1,.5)$ & .18 (9.89) & .10 (15.16) & .063 (18.17) & .20 (71.26) & .411 (240.53) & .375 (298.27) & .42 (1337.00) \\
    {} & $(5,.5)$ & .65 (7.55) & .87 (11.94) & .180 (18.29) & .70 (60.56) & .694 (198.40) & .723 (282.02) & .62 (767.81) \\
    \midrule
    \multirow[t]{2}{*}{mbGW} & $(50,5K)$ & .22 (19.17) & .44 (17.95) & .043 (19.90) & .74 (22.13) & .494 (25.88) & .325 (26.60) & .51 (31.51) \\
    {} & $(50,0.1)$  & .30 (.71)  & .61 (.72) & .048 (1.02) & .78 (2.23) & .506 (4.51) & .334 (5.35) & .52 (10.18) \\
    \midrule
    \multirow[t]{2}{*}{qGW} & $.01$ & .36 (.06) & .53 (.09) & .044 (.13) & \underline{ .18 (.34)} & .330 (.93) & .161 (1.22) & .26 (5.96) \\
    {} & $.1$ & .28 (.32)  & .08 (.70) & \underline{.016 (.78)} & .28 (1.48) & .002 (4.20) & \underline{.026 (5.96)} & \underline{{\bf .18 (25.23)}} \\
    {} & $.2$ & \underline{.14 (.70)} & \underline{{\bf .03 (1.15)}} & .020 (1.56) & .22 (4.10) & \underline{{\bf .001 (11.37)}} &  .002 (15.38) & .21 (74.43) \\
    {} & $.5$ & .06 (2.71) & .11 (4.82) & {\bf .010 (6.94)} & .19 (26.02) & .001 (89.82) & {\bf .001 (122.26)} & .21 (642.09) \\
    \bottomrule
    \end{tabular}}
\end{table*}

\paragraph{Graph Matching.}
\label{sec:exp-graph-match}

\begin{table*}
    \centering
    \caption{Distortion percentage (lower is better) and runtimes (s) on graph matching. Blank entries correspond to experiments that did not complete in 1 hour or ran out of memory for storing distance matrices.}
    \label{tab:graph-matching}
    \resizebox{\textwidth}{!}{
    \begin{tabular}{
    p{0.09\textwidth} p{0.12\textwidth} 
    P{0.15 \textwidth} P{0.15\textwidth}
    P{0.15\textwidth} P{0.15\textwidth}
    P{0.16\textwidth} P{0.18\textwidth} P{0.15\textwidth}}
    \toprule 
    Method & Param &
    Centaur 1 & Centaur 2 & Centaur 3 & Centaur 4
    & Centaur 5 & Cat & David \\
    {} & {} &
    15768 & 15768 & 15768 & 15768
    & 15768 & 27894 & 52565 \\
    \midrule
    \multirow[t]{4}{*}{erGW} & $10^3$ & {92.2 (1059)} & 92.0 (1080) & 91.7 (1074)  & { 91.7 (1060)} & 92.0 (1054) & --- & --- \\
    \midrule
    \multirow[t]{2}{*}{mbGW} & $(400,2K)$ & 48.3 (788) & 48.1 (784) & 46.3 (778) & 46.4 (783) & 48.0 (779) & 42.9 (712) & --- \\
    \midrule
    \multirow[t]{2}{*}{MREC} & $(750,10^{-3})$ & 78.6 (273) & 13.3 (267) & 14.5 (288) & 13.8 (255) & 22.7 (272) & 70.1 (707) & --- \\
    \midrule
    \multirow[t]{2}{*}{qFGW} & $(0.5,0.75)$ & {\bf6.76} (4.53) & {\bf6.62} (4.54) & {\bf6.65} (4.56) & {\bf6.55} (4.52) & {\bf6.71} (4.68) & {\bf8.28} (7.83) & {\bf82.5} (8.62) \\
    \bottomrule
    \end{tabular}}
\end{table*}

Graph matching is a fundamental GW application for which \cite{xu2019scalable} produced state-of-the-art scalability results, including matching source and target graphs with 2K and 9K nodes, respectively. Here we match graphs coming from meshes in the TOSCA dataset \cite{bronstein2008numerical}. We choose multiple models (number of meshes in parentheses) from the ``Centaur" (6), ``Cat" (2), and ``David" (2) mesh families, having approximately 16K, 30K, and 50K vertices, respectively. Meshes in each category correspond to different poses of the same object, and the underlying vertices are numbered in a compatible way to provide for ground truth labels. For the six Centaur graphs, we wished to compute GW matchings between $(G_1,G_2),\ldots, (G_5,G_6)$. We retained the structure of the Point Cloud Matching experiment with three variations. First, for evaluating the quality of a matching, we computed the ratio of the summed distortions of a matching $\mu$ (graph distance from each matched point to its ground truth copy) to the distortion of a \emph{random} matching (averaged over five random matchings), and converted this into a percentage. Thus a lower distortion score is better, and these are the scores reported in Table \ref{tab:graph-matching}. Second, we used the observation of \cite{vayer2019optimal} that adding node features via Weisfeiler-Lehman (WL) leads to superior performance, and devised a WL scheme to apply qFGW. Third, to demonstrate how to use these methods in a cross-validation pipeline, we took the two fastest methods---qFGW and MREC---and optimized parameters for them using leave-one-out cross-validation on the Centaur comparisons. For MREC this cross-validation produced $\epsilon=10^{-3}$ and \#clusters=750, and for qFGW we fixed $m=1000$ and obtained $\alpha=0.5,\beta=0.75$. Results on the five testing folds are presented in Table \ref{tab:graph-matching}; these optimized parameters are also used for the Cat and David comparisons.

\paragraph{Application to Segmentation Transfer.} 
\label{sec:exp-segment-transfer}

\begin{figure}
    \centering
    \includegraphics[width=\textwidth]{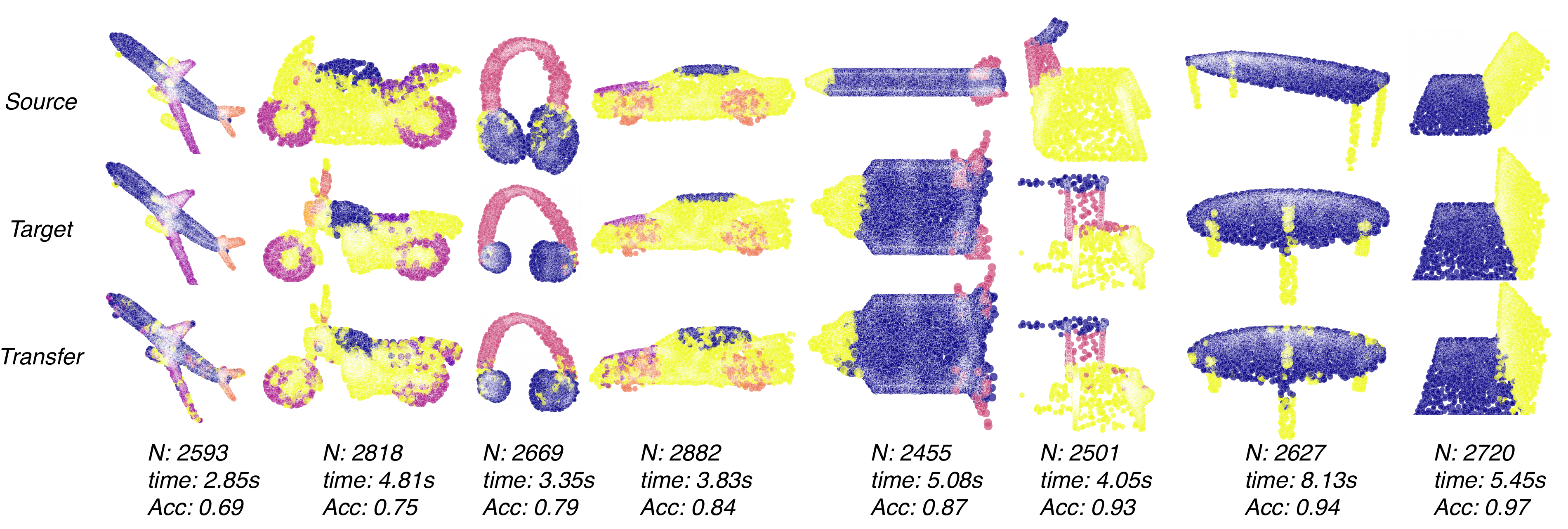}
    \caption{Semantic segmentation transfer on ShapeNet. Colors denote part annotations.}
    \label{fig:seg-transfer}
\end{figure}

Given point cloud datasets with semantic labels, segmentation transfer is the problem of constructing correspondences that preserve segment category labels. We demonstrate the applicability of qFGW to segmentation transfer using the ShapeNet CAD model dataset. This dataset contains 16 categories, each category having objects with approximately 3K points split into 2-6 parts. Point features are chosen to be surface normals. We choose 12 models each from the eight categories $\{$Airplane, Car, Earphone, Guitar, Laptop, Motorbike, Rocket, Table$\}$. We optimize parameters over a simple grid of $\alpha,\beta$ parameters as in the Graph Matching experiment, and illustrate results with optimal parameters in Figure \ref{fig:seg-transfer}. For evaluation, we obtain matchings $\mu$ via an argmax as in the preceding experiments, and then count the fraction of matches between source and target part labels normalized by the number of points.

\paragraph{Application to Large Scale Segment Transfer.}

\label{sec:exp-large-matching}
\begin{figure}
    \centering
    \includegraphics[width=\textwidth]{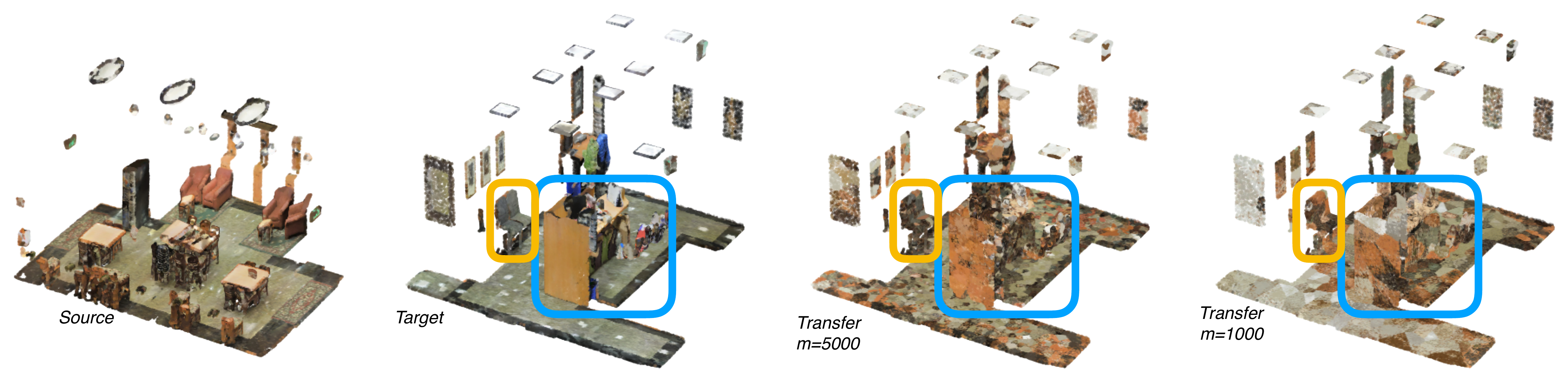}
    \caption{Partial render of Lobby rooms ($\sim$1M points) in the S3DIS dataset. Note that the target room has furniture of different types than the source room. Boxes show improvement in segment transfer to the chair and desk from using more landmarks.}
    \label{fig:large-matching}
\end{figure}

As further demonstration of segment transfer on extremely large metric spaces, we use the Stanford 3D Indoor Scene Dataset (S3DIS) \cite{armeni20163d}. This point cloud dataset comprises six areas containing 271 rooms, totaling to 215 million points where each point is labeled with one of 13 semantic categories and comes with an RGB color vector. We choose two Lobby rooms in Area 4, one containing 1,155,072 points and the other containing 909,312 points. We carry out matching using qFGW, using point colors as features. For evaluation we obtain a matching $\mu$, count the number of points that get matched to a point of the same part category, and normalize by the number of points in the source room. Because the rooms may contain parts belonging to different semantic categories, making direct evaluation difficult, we compare against a random matching (higher is better): random matching obtains 10.0\%, matching with $m=1000$ obtains 26.2\%, and matching with $m=5000$ obtains 41.0\%. Notably, the overall computation is completed in just 10 minutes (for $m=1000$) on a standard Macbook Pro with 8GB RAM.

\section{Discussion}

By combining the crucial insight of partitioning for scalable GW \cite{blumberg2020mrec,xu2019scalable}, the notion of slicing a mm-space by distance to anchor points \cite{dghlp-focm,vayer2019sliced} and ideas regarding the duality between quantizing and clustering a metric space \cite{memoli2018sketching}, we presented qGW: a theoretical framework for scalable GW computation with error bounds as well as new algorithmic improvements in time and space complexity. These error bounds complement those stated in \cite{blumberg2020mrec} via the language of \emph{doubling dimension} in metric spaces, and we related these notions using the concept of quantized eccentricity and associated bounds. Future theoretical work could validate this dependence on dimension by examining other intrinsic properties that could potentially tighten these bounds. This in turn motivates us to study how to extract local structure to aid with GW alignment even in high dimensional spaces.

\section{Acknowledgements}

We would like to thank Mathieu Carri\`{e}re for help with the MREC code, Vikas Garg for sharing code from \cite{garg2019solving} and Facundo M\'{e}moli for providing useful  feedback.

\bibliographystyle{splncs04}
\bibliography{biblio}

\pagebreak

\appendix

\section{Proofs}

\subsection{Proof of Proposition \ref{prop:quantization_couplings}}

Letting $\pi_X:X \times Y \to X$ and $\pi_Y:X \times Y \to Y$ denote the coordinate projection maps, the marginal condition on a coupling $\mu \in \mathcal{C}(\mu_X,\mu_Y)$ can be expressed in terms of \emph{pushforward measures} as $(\pi_X)_\# \mu = \mu_X$ and $(\pi_Y)_\# \mu = \mu_Y$. Let $\mu$ be a quantization coupling of the form \eqref{eqn:quantization_coupling}. Let $A \subset X$ and consider $(\pi_X)_\# \mu(A)$. Then we have
    \begin{align}
        (\pi_X)_\# \mu(A) &= \mu\left(\pi_X^{-1}(A)\right) = \sum_{p,q} \bar{\mu}_{x^p,y^q} \left(\pi_X^{-1}(A)\right) \mu_m(x^p,y^q) \nonumber \\
        &= \sum_{p,q} \bar{\mu}_{U^p}(A) \mu_m(x^p,y^q) = \sum_{p,q} \mu_{U^p}(A \cap U^p) \mu_m(x^p,y^q) \nonumber \\
        &= \sum_p \frac{1}{\mu_X(U^p)}\mu_{X}(A \cap U^p) \sum_q \mu_m(x^p,y^q)  \nonumber \\
        &= \sum_p \frac{1}{\mu_X(U^p)}\mu_{X}(A \cap U^p) \mu_{\mcP_X}(x^p)   \label{eqn:quantization_coupling_1}\\
        &= \sum_p \mu_X(A \cap U^p) \label{eqn:quantization_coupling_2} \\
        &= \mu_X(A), \nonumber
    \end{align}
where \eqref{eqn:quantization_coupling_1} follows by the marginal condition on $\mu_m$ and \eqref{eqn:quantization_coupling_2} follows because $\mu_{\mcP_X}(x^p) = \mu_X(U^p)$. The marginal condition $\left(\pi_Y\right)_\# \mu = \mu_Y$ follows similarly.

\subsection{Full proof of Theorem \ref{thm:qGW_is_a_metric}}

Symmetry is obvious. If the pointed mm-spaces are isomorphic, then quantized GW distance between them is zero distance, as this value is realized by the coupling induced by the isomorphism. On the other hand, if their distance is zero then we must show that a they are isomorphic as pointed mm-spaces. We first observe that $\dqgw$ is always realized by a quantization coupling. Indeed, this follows because the Gromov-Wasserstein loss $\mathrm{GW}$ is continuous and the set of quantization couplings is compact (in the Euclidean topology). To see this last point, observe that each set $\mathcal{C}(\mu_{\mcP_X},\mu_{\mcP_Y})$ and $\mathcal{C}(\overline{\mu}_{U^p},\overline{\mu}_{V^q})$ is closed (e.g., \cite[p. 49]{villani2003topics}), and this implies that a convergent sequence of quantization couplings must converge to a quantization coupling. Closedness follows and boundedness is obvious. Now suppose that $\dqgw((X,\mcP_X),(Y,\mcP_Y)) = 0$ and choose a quantization coupling $\mu$ whose GW loss is zero. By \cite[Lemma 10.4]{dghlp-focm}, the support of $\mu$ is of the form $\{(x,\phi(x)) \mid x \in X\}$, where $\phi:X \to Y$ is a measure-preserving isometry. Then for each $x^r \in X^m$, we have
\[
0 < \mu(x^r,\phi(x^r)) = \sum_{p,q} \mu_m(x^p,y^q)\overline{\mu}_{x^p,y^q}(x^r,\phi(x^r)).
\]
For $r \neq p$, we have $\overline{\mu}_{x^p,y^q}(x^r,\cdot) = 0$. Furthermore, $\phi(x^r)$ lies in exactly one partition block $V^q$. From this, we conclude that
\[
0 < \mu_m(x^r,y^q)\overline{\mu}_{x^r,y^q}(x^r,\phi(x^r)).
\]
Since we assumed that quantization couplings have the property that $\mu_{x^r,y^q}(x^r,y^q) > 0$, it must be that $\phi(x^r) = y^q$. This proves that the mm-space isomorphism $\phi$ is actually an isomorphism of pointed mm-spaces.

It remains to prove that the triangle inequality holds. Let $(X,\mcP_X)$, $(Y,\mcP_Y)$ and $(Z,\mcP_Z)$ be $m$-pointed finite mm-spaces. For the sake of simplifying notation in the proof, assume that $\mu_X$, $\mu_Y$ and $\mu_Z$ are fully supported. Let $\mu^{XY}$ and $\mu^{YZ}$ be quantization couplings of $X$ with $Y$ and $Y$ with $Z$, respectively. We construct a coupling $\mu^{XZ}$ of $X$ and $Z$ by the formula
\[
\mu^{XZ}(x,z) := \sum_{y \in Y} \frac{\mu^{XY}(x,y)\mu^{YZ}(y,z)}{\mu_Y(y)}.
\]
Unwrapping definitions in the finite setting, this is the same coupling that is used to demonstrate the triangle inequality for standard GW distance \cite[Theorem 5.1]{dghlp-focm}, which is based on an application of the gluing lemma from optimal transport theory \cite[Lemma 7.6]{villani2003topics}. It follows that 
\[
\mathrm{GW}(\mu^{XZ})^{1/2} \leq \mathrm{GW}(\mu^{XY})^{1/2} + \mathrm{GW}(\mu^{YZ})^{1/2}.
\]
To finish the proof, it suffices to show that this construction of $\mu^{XZ}$ yields a quantization coupling, which is an exercise in manipulating summation notation. We have
\begin{align}
    &\sum_y \frac{\mu^{XY}(x,y)\mu^{YZ}(y,z)}{\mu_Y(y)} \nonumber \\
    &\qquad = \sum_y \frac{1}{\mu_Y(y)} \left(\sum_{p,q} \mu_m^{XY}(x^p,y^q)\overline{\mu}_{x^p,y^q}(x,y)\right) \nonumber \\
    &\hspace{1.75in} \cdot
    \left(\sum_{q,r} \mu_m^{YZ}(y^q,z^r)\overline{\mu}_{y^q,z^r}(y,z)\right) \nonumber \\
    &\qquad = \sum_y \frac{1}{\mu_Y(y)} \sum_{p,q,q',r} \mu_m^{XY}(x^p,y^q) \mu_m^{YZ}(y^{q'},z^r) \overline{\mu}_{x^p,y^q}(x,y) \overline{\mu}_{y^{q'},z^r}(y,z) \label{eqn:cgw_proof_1} \\
    &\qquad = \sum_y \frac{1}{\mu_Y(y)} \sum_{p,q,r} \mu_m^{XY}(x^p,y^q) \mu_m^{YZ}(y^{q},z^r) \overline{\mu}_{x^p,y^q}(x,y) \overline{\mu}_{y^{q},z^r}(y,z) \label{eqn:cgw_proof_2},
\end{align}
where the prime notation $q'$ in \eqref{eqn:cgw_proof_1} is introduced to indicate that partition representatives $y^q$ and $y^{q'}$ are being summed over independently. This notation is removed in \eqref{eqn:cgw_proof_2}; this is allowed by the observation that the factor $\overline{\mu}_{x^p,y^q}(x,y) \overline{\mu}_{y^{q'},z^r}(y,z)$ in \eqref{eqn:cgw_proof_1} is nonzero only when the points lie in partition blocks $x \in U^p$, $y \in V^q$, $y \in V^{q'}$ and $z \in W^r$, which implies $q = q'$. Continuing, the above is equal to
\begin{align}  
    &\qquad = \sum_y \frac{1}{\mu_Y(y)}  \sum_{p,q,r} \frac{\mu_m^{XY}(x^p,y^q) \mu_m^{YZ}(y^{q},z^r)}{\mu_{\mcP_Y}(y^q)} \mu_{\mcP_Y}(y^q) \overline{\mu}_{x^p,y^q}(x,y) \overline{\mu}_{y^{q},z^r}(y,z) \nonumber \\
    & \qquad = \sum_y \frac{1}{\mu_Y(y)}  \left(\sum_{p,q,r} \frac{\mu_m^{XY}(x^p,y^q) \mu_m^{YZ}(y^{q},z^r)}{\mu_{\mcP_Y}(y^q)}\right)  \nonumber \\
    &\hspace{1.75in} \cdot \left(\sum_q \mu_{\mcP_Y}(y^q) \overline{\mu}_{x^p,y^q}(x,y) \overline{\mu}_{y^{q},z^r}(y,z) \right) \label{eqn:cgw_proof_3} \\
    &\qquad = \sum_{p,r} \left(\sum_{q} \frac{\mu_m^{XY}(x^p,y^q) \mu_m^{YZ}(y^{q},z^r)}{\mu_{\mcP_Y}(y^q)}\right) \nonumber \\
    &\hspace{1.75in} \cdot \left(\sum_q \mu_{\mcP_Y}(y^q) \sum_y \frac{\overline{\mu}_{x^p,y^q}(x,y) \overline{\mu}_{y^{q},z^r}(y,z)}{\mu_Y(y)}\right), \label{eqn:cgw_proof_4}
\end{align}
where line \eqref{eqn:cgw_proof_3} follows because only one of the terms in the second sum over $q$ indices is nonzero. The last equality in \eqref{eqn:cgw_proof_4} follows by rearranging summations.

To complete the proof, we observe that \eqref{eqn:cgw_proof_4} gives the desired conclusion. This is because the term
\[
\sum_{q} \frac{\mu_m^{XY}(x^p,y^q) \mu_m^{YZ}(y^{q},z^r)}{\mu_{\mcP_Y}(y^q)}
\]
is a coupling of $\mu_{\mcP_X}$ and $\mu_{\mcP_Z}$ (it is the coupling of these measures obtained by the Gluing Lemma). Likewise, each 
\[
\sum_y \frac{\overline{\mu}_{x^p,y^q}(x,y) \overline{\mu}_{y^{q},z^r}(y,z)}{\mu_Y(y)}
\]
is a coupling of $\overline{\mu}_{U^p}$ and $\overline{\mu}_{W^r}$, where $U^p$ and $W^r$ are the partition blocks represented by $x^p$ and $z^r$, respectively. Since $\sum_q \mu_{\mathcal{P}_Y}(y^q) = 1$ and each term in the sum is positive, convexity of $\mathcal{C}(\mu_{U^p},\mu_{W^r})$ implies that 
\[
\sum_q \mu_{\mcP_Y}(y^q) \sum_y \frac{\overline{\mu}_{x^p,y^q}(x,y) \overline{\mu}_{y^{q},z^r}(y,z)}{\mu_Y(y)} \in \mathcal{C}(\mu_{U^p},\mu_{W^r}).
\]
Therefore $\mu$ is a quantization coupling.

\subsection{Proof of Proposition \ref{prop:locally_linear_solution}}

The main statement follows from \cite[Lemma 27]{gwnets}. The idea is that we write $d_X(x_i,x^p) =: r_i \in \R$ and $d_Y(y_j,y^q) =: s_j \in \R$ for each $x_i \in U^p$ and $y_j \in V^q$. Then \eqref{eqn:locally_linear} is equivalent to solving
\[
\min_{\nu \in \mathcal{C}(\nu_{U^p},\nu_{V^q})} \sum_{i,j} (r_i - s_j)^2 \mu(r_i,s_j),
\]
where
\[
\nu_{U^p} := d_X(\cdot,x^p)_\# \mu_{U^p} \quad \mbox{ and } \quad \nu_{V^q} := d_Y(\cdot,y^q)_\# \mu_{V^q}
\]
are pushforward measures on the real line. It is well known that optimal transport on the real line can be solved efficiently \cite[Section 2.6]{peyre2019computational}---the $O(k\log(k))$ complexity is due to the need to sort points in each partition block by distance to the anchor.

\subsection{Proof of Lemma \ref{lem:quantized eccentricity_bound}}

Let $\mcP_X = \{(x^1,U^1),\ldots,(x^m,U^m)\}$ be an $m$-pointed partition of $X$ and let $X^m = \{x^1,\ldots,x^m\}$. Recall that $\mu_{\mcP_X}$ is a probability measure on $X^m$ with $\mu_{\mcP_X}(x^p) = \mu_X(U^p)$. Each $U^p$ is also considered as a mm-space, with measure $\mu_{U^p}$ as defined in Section \ref{sec:qGW_metric}. Define a measure $\mu$ on $X \times X^m$ by 
\[
\mu(x_i,x^p) := \left\{\begin{array}{cc}
\mu_X(x) & \mbox{ if $x \in U^p$}\\
0 & \mbox{ otherwise.}\end{array}\right.
\]
It is easy to check that $\mu \in \mathcal{C}(\mu_X,\mu_{\mathcal{P}_X})$. Then we have
\begin{align}
    \dgw(X,X^m) &\leq \left(\sum_{i,p,j,q} \left(d_X(x_i,x_j) - d_X(x^p,x^q)\right)^2 d\mu(x_i,x^p)d\mu(x_j,x^q)\right)^{1/2} \nonumber \\
    &\leq \left(\sum_{i,p} \left(d_X(x_i,x^p)\right)^2 \mu(x_i,x^p) \right)^{1/2} \nonumber \\
    &\qquad \qquad + \left(\sum_{j,q} \left(d_X(x_j,x^q)\right)^2 \mu(x_j,x^q) \right)^{1/2} \label{eqn:quant_ecc_1} \\
    &\leq \left(\sum_{p} \sum_{x_i \in U^p} \left(d_X(x_i,x^p)\right)^2 \mu(x_i,x^p) \right)^{1/2} \nonumber \\
    &\qquad \qquad + \left(\sum_{q} \sum_{x_j \in U^q} \left(d_X(x_j,x^q)\right)^2 \mu(x_j,x^q) \right)^{1/2} \label{eqn:quant_ecc_2} \\
    &\leq \left(\sum_{p} \mu_X(U^p) \sum_{x_i \in U^p} \left(d_X(x_i,x^p)\right)^2 \mu_{U^p}(x_i) \right)^{1/2} \nonumber \\
    &\qquad \qquad + \left(\sum_{q} \mu_X(U^q) \sum_{x_j \in U^q} \left(d_X(x_j,x^q)\right)^2 \mu_{U^q}(x_j) \right)^{1/2} \label{eqn:quant_ecc_3} \\
    &= 2 \left(\sum_p \mu_X(U^p) s_{U^p}(x^p)^2\right)^{1/2} = 2q(\mcP_X), \nonumber
\end{align}
which proves the main claim of the lemma. The second claim follows by minimizing both sides of the inequality over $m$-pointed partitions. The steps in the chain of inqualities are justified as follows: Equation \eqref{eqn:quant_ecc_1} follows by an application of the triangle inequality on $d_X$, Minkowski's inequality on $\|\cdot\|_{L^2(\mu \otimes \mu)}$ and the fact that  $\mu$ is a probability measure; Equation \eqref{eqn:quant_ecc_2} follows by the definition of $\mu$; Equation \eqref{eqn:quant_ecc_3} follows by the definition of $\mu$ and the definitions of $\mu_{U^p}$ and $\mu_{U^q}$.

\subsection{Proof of Theorem \ref{thm:delta_bound}}

Let $X$ and $Y$ be finite mm-spaces with $m$-pointed partitions 
\[
\mcP_X = \{(x^1,U^1)\,\ldots,(x^m,U^m)\} \quad \mbox{and} \quad \mcP_Y = \{(y^1,V^1),\ldots,(y^m,V^m)\},
\]
respectively. As usual, we denote by $X^m$ and $Y^m$ the respective sets of partition blocks. We consider each $X^m$, $U^p$, $Y^m$, $V^q$ as a mm-space in its own right, using the notation and terminology of Section \ref{sec:qGW_metric}. Let $p_X:X \to \X^m$ and $p_Y:Y \to Y^m$ denote the obvious projection maps; i.e., $p_X:x \mapsto x^p$ when $x \in U^p$. 

Let $x \in U^p, x' \in U^{p'}$, $y \in V^q, y' \in V^{q'}$. We define
\begin{align*} 
J(x,y,x',y')&:= d_X(x,x') - d_Y(y,y')\\
J^*(x,y,x',y')&:= d_X(x,p_X(x)) +d_X(p_X(x),p_X(x')) + d_X(p_X(x'),x') \\
&\qquad - d_Y(y,p_Y(y)) -d_Y(p_Y(y),p_Y(y')) - d_Y(p_Y(y'),y')\\
&=d_X(x,x^p) +d_X(x^p,x^{p'}) + d_X(x^{p'},x') \\
&\qquad \qquad - d_Y(y,y^q) - d_Y(y^q,y^{q'}) - d_Y(y^{q'},y').
\end{align*}
Assuming that the partition blocks of $\mcP_X$ and $\mcP_Y$ have metric diameter less than $\epsilon$, we have
\begin{align*}
&| J(x,y,x',y') - J^*(x,y,x',y')| \\
& \qquad \leq |d_X(x,x') - d_X(x^p,x^{p'})| + |d_Y(y,y') - d_Y(y^q,y^{q'})| \\
& \qquad \qquad + |d_X(x,x^p) - d_Y(y,y^q)| + |d_X(x^{p'},x') - d_Y(y^{q'},y')| \leq 6\e.
\end{align*}
Now observe that the Gromov-Wasserstein loss of a coupling $\mu$ is given by 
\[
\mathrm{GW}(\mu) = \|J\|_{L^2(\mu \otimes \mu)}^2.
\]
We similarly define a loss $\mathrm{GW}^\ast$ on $\mathcal{C}(\mu_X,\mu_Y)$ by
\[
\mathrm{GW}^\ast(\mu) := \|J^\ast\|_{L^2(\mu \otimes \mu)}^2.
\]
By Minkowski's inequality and the estimate above, we have
\begin{align}
    \left|\mathrm{GW}(\mu)^{1/2} - \mathrm{GW}^\ast(\mu)^{1/2}\right| &= \left| \| J \|_{L^2(\mu \otimes \mu)} - \| J^* \|_{L^2(\mu \otimes \mu)} \right| \nonumber \\
    &\leq \| J - J^* \|_{L^2(\mu \otimes \mu)}  \leq 6\epsilon \label{eqn:delta_bound_11}.
\end{align}

We proceed by estimating the loss $\mathrm{GW}^\ast$ of a coupling obtained via the qGW algorithm. Let $\mu = \sum_{p,q} \mu_m(x^p,y^q) \mu_{x^p,y^q}$ be a coupling obtained via the algorithm: $\mu_m$ is an optimal couplign of $X^m$ and $Y^m$ and each $\mu_{x^p,y^q}$ solves the local linear matching problem \eqref{eqn:locally_linear}. Then we have that $\mathrm{GW}^\ast(\mu)$ is equal to
\begin{align*}
    & \sum_{i,j,k,\ell} \left(J^\ast(x_i,y_j,x_k,y_\ell)\right)^2 \mu(x_i,y_j) \mu(x_k,y_\ell) \\
    &= \sum_{i,j,k,\ell} \left(d_X(x_i,p_X(x_i)) +d_X(p_X(x_i),p_X(x_k)) + d_X(p_X(x_k),x_k)\right. \\
&\qquad \left.- d_Y(y_j,p_Y(y_j)) -d_Y(p_Y(y_j),p_Y(y_\ell)) - d_Y(p_Y(y_\ell),y_\ell)\right)^2\mu(x_i,y_j) \mu(x_k,y_\ell).
\end{align*}
Now we define $f:= (p_X,p_Y):X \times Y \to X^m \times Y^m$ and let 
\[
\widehat{\mu}:=(\mathrm{id}_{X\times Y},f)_\# \mu \in \mathcal{C}(\mu_X \otimes \mu_Y, \mu_{\mcP_X} \otimes \mu_{\mcP_Y}).
\]
Using marginalization constraints, the expression for $\mathrm{GW}^\ast(\mu)$ given above can be expressed as \begin{align*}
    &\sum_{i,j,k,\ell,p,q,p',q'} \left(d_X(x_i,p_X(x_i)) +d_X(p_X(x_i),p_X(x_k)) + d_X(p_X(x_k),x_k)\right. \\
&\hspace{1in} \left.- d_Y(y_j,p_Y(y_j)) -d_Y(p_Y(y_j),p_Y(y_\ell)) - d_Y(p_Y(y_\ell),y_\ell)\right)^2 \\
&\hspace{2.5in} \cdot \widehat{\mu}(x_i,y_j,x^p,y^q) \widehat{\mu}(x_k,y_\ell,x^{p'},y^{q'}).
\end{align*}
Since $\widehat{\mu}$ is supported on tuples $(x,y,x^p,y^q)$ such that $p_X(x) = x^p$ and $p_Y(y) = y^q$, the above can be expressed as 
\begin{align}
    &\sum_{i,j,k,\ell,p,q,p',q'} \left(d_X(x_i,x^p) +d_X(x^p,x^{p'}) + d_X(x^{p'},x_k)\right. \nonumber \\
&\hspace{1in} \left.- d_Y(y_j,y^q) -d_Y(y^q,y^{q'}) - d_Y(y^{q'},y_\ell)\right)^2 \widehat{\mu}_{ijpq} \widehat{\mu}_{k,\ell,p',q'}, \nonumber 
\end{align}
where $\widehat{\mu}_{ijpq} = \widehat{\mu}(x_i,y_j,x^p,y^q)$ is shorthand notation. 
In turn, we deduce by Minknowski's inequality that
\begin{align*}
    &\mathrm{GW}^\ast(\mu)^{1/2} \\
    &\leq \left(\sum_{i,j,k,\ell,p,q,p',q'} \left(d_X(x_i,x^p) - d_Y(y_j,y^q)\right)^2 \widehat{\mu}_{ijpq} \widehat{\mu}_{k,\ell,p',q'}\right)^{1/2} \\
    &\qquad + \left(\sum_{i,j,k,\ell,p,q,p',q'} \left(d_X(x_k,x^{p'}) - d_Y(y_\ell,y^{q'})\right)^2 \widehat{\mu}_{ijpq} \widehat{\mu}_{k,\ell,p',q'}\right)^{1/2} \\
    &\qquad + \left(\sum_{i,j,k,\ell,p,q,p',q'} \left(d_X(x^p,x^{p'}) - d_Y(y^q,y^{q'})\right)^2 \widehat{\mu}_{ijpq} \widehat{\mu}_{k,\ell,p',q'}\right)^{1/2}.
\end{align*}
Using our diameter bound and the fact that $\widehat{\mu}$ is a probability measure, each of the first two terms in the above is bounded above by $\epsilon$. By the marginal constraints of $\widehat{\mu}$ and the assumption that $\mu_m$ was an optimal coupling of $X^m$ awith $Y^m$, the last term is equal to the GW distance between $X^m$ and $Y^m$. We deduce that
\begin{equation}\label{eqn:delta_bound_22}
\mathrm{GW}^\ast(\mu)^{1/2} \leq 2\epsilon + \dgw(X^m,Y^m).
\end{equation}

We now complete the proof of the theorem. We have
\begin{align}
    &\left|\dgw(X,Y) - \delta((X,\mcP_X),(Y,\mcP_Y)) \right| \nonumber \\
    &\hspace{1.75in} = \delta((X,\mcP_X),(Y,\mcP_Y)) - \dgw(X,Y) \label{eqn:delta_bound_1} \\
    &\hspace{1.75in} = \mathrm{GW}(\mu)^{1/2} - \dgw(X,Y)  \label{eqn:delta_bound_2}\\
    &\hspace{1.75in} \leq \mathrm{GW}^\ast(\mu)^{1/2} + 6 \epsilon - \dgw(X,Y) \label{eqn:delta_bound_3}\\
    &\hspace{1.75in} \leq 2 \epsilon  + \dgw(X^m,Y^m) + 6 \epsilon - \dgw(X,Y) \label{eqn:delta_bound_4}\\
    &\hspace{1.75in} \leq 2(q(\mathcal{P}_X) + q(\mathcal{P}_Y)) + 8 \epsilon \label{eqn:delta_bound_5}.
\end{align}
The equality \eqref{eqn:delta_bound_1} follows because GW distance minimizes over a larger set of couplings than $\delta$ and the equality \eqref{eqn:delta_bound_2} follows by the assumption that our $\mu$ is the coupling obtained through the qGW algorithm. Inequality \eqref{eqn:delta_bound_3} follows by \eqref{eqn:delta_bound_11}, \eqref{eqn:delta_bound_4} follows by \eqref{eqn:delta_bound_22} and \eqref{eqn:delta_bound_5} follows by collecting terms and by (the proof of) Theorem \ref{thm:gw_estimation}.

\section{A Simple Comparison Against GW}\label{sec:baseline_gw}

Since qGW attempts to minimize GW loss \eqref{eqn:gw_loss} over a restricted set of couplings, a simple way to ascertain the quality of matchings obtained via qGW is to compare their GW loss to the GW loss of a coupling obtained via the standard GW algorithm. This simple baseline is explored in the following experiment. We generate pairs $X,Y$ of random planar pointclouds of size $N \in \{200,400,\ldots,2000\}$, each created via the \texttt{make\underline{ }blobs} function from the \texttt{scikit-learn} package \cite{scikit-learn}. Each point cloud is treated as an mm-space with Euclidean distance and uniform measure. For each pair we compute a GW coupling and a qGW coupling. Specifically, qGW is computed by random Voronoi partitions at various levels of sampling $\{.1,.2.3,.4,.5\}$, similar to what was done in Section \ref{sec:point_cloud_matching}. The \emph{relative error} of the qGW coupling $\mu_{qGW}$ is 
\[
\frac{\mathrm{GW}(\mu_{prod}) - \mathrm{GW}(\mu_{qGW})}{\mathrm{GW}(\mu_{prod}) - \mathrm{GW}(\mu_{GW})},
\]
where $\mu_{prod}$ is the product coupling and $\mu_{GW}$ is the standard GW coupling. The idea is that $\mu_{GW}$ is the putative minimum of the GW loss (although it is really only an approximation of the minimum, due to computational limitations) and $\mu_{prod}$ is the treated as the putative maximum. 

\begin{figure}
\includegraphics[width=\textwidth]{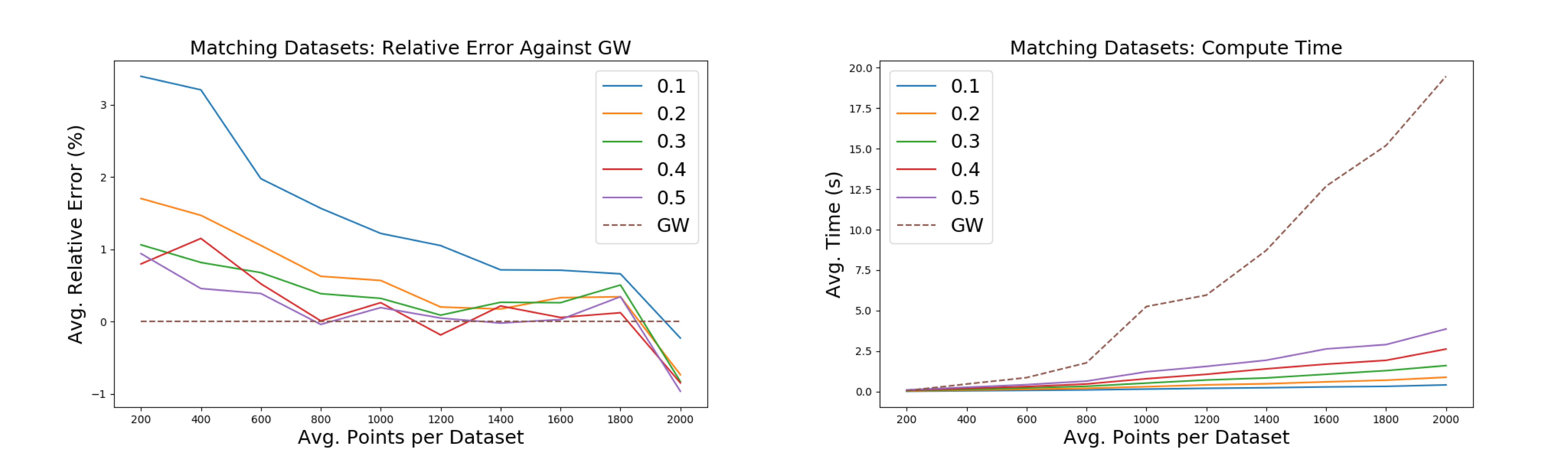}
\caption{GW baseline comparison. Left shows the relative error (see text) of the qGW algorithm (for various levels of subsampling) against GW in matching ``blobs" datasets with an increasing number of points. Right shows the average compute time per match for GW and qGW at various levels of subsampling.} \label{fig:matching_blobs}
\end{figure}

For each $N$, we produce $10$ pairs of point clouds, compute the relative error of the qGW coupling between them and average the results. These average relative errors are reported across different values of $N$ in Figure \ref{fig:matching_blobs}. Observe that the relative error is sometimes negative, indicating that qGW found a \emph{better} local minimizer of GW loss than the standard GW algorithm did. We also report the average times to compute the matchings for each method. 

\end{document}